\documentclass[lettersize,journal]{IEEEtran}

\usepackage[utf8]{inputenc} % allow utf-8 input
\usepackage[T1]{fontenc}    % use 8-bit T1 fonts
\usepackage{hyperref}       % hyperlinks
\usepackage{url}            % simple URL typesetting
\usepackage{booktabs}       % professional-quality tables
\usepackage{amsfonts}       % blackboard math symbols
\usepackage{amsmath}
\usepackage{nicefrac}       % compact symbols for 1/2, etc.
\usepackage{microtype}      % microtypography
\usepackage{verbatim}       % verbatim, for comments
\usepackage{float}
\usepackage{subcaption}
\usepackage{multirow}
\usepackage{caption}
\usepackage{graphicx}
\usepackage[table,xcdraw]{xcolor}
\usepackage{dirtree}
\usepackage[procnumbered,ruled]{algorithm2e} % This adds a space between the paragraphs, it can be removed later if need be, but it's hard to read the paragraphs while working when all squished together.
\SetKwInput{Parameter}{Parameter}
\SetKwInput{Input}{Input}
\SetKwInput{Output}{Output}
\newcommand{\TheName}{\textbf{GLARE}}
\usepackage{tabularx}
\usepackage{enumitem}

\title{\LARGE \bf $\TheName$: A Dataset for Traffic Sign Detection in Sun Glare
}

\author{\IEEEauthorblockN{
Nicholas Gray\IEEEauthorrefmark{1}\IEEEauthorrefmark{5},
Megan Moraes\IEEEauthorrefmark{1}\IEEEauthorrefmark{5},
Jiang Bian\IEEEauthorrefmark{1}\IEEEauthorrefmark{5}\IEEEauthorrefmark{3},
Alex Wang\IEEEauthorrefmark{1}\IEEEauthorrefmark{4},\IEEEauthorrefmark{6},\\
Allen Tian\IEEEauthorrefmark{4} \IEEEauthorrefmark{7},
Kurt Wilson\IEEEauthorrefmark{5}\IEEEauthorrefmark{2},
Yan Huang\IEEEauthorrefmark{5},
Haoyi Xiong\IEEEauthorrefmark{3},
Zhishan Guo\IEEEauthorrefmark{5}\IEEEauthorrefmark{2}
\\
\IEEEauthorrefmark{5}University of Central Florida, \IEEEauthorrefmark{3}Baidu Research Lab , \IEEEauthorrefmark{4}Local High Schools, 
\IEEEauthorrefmark{6}Georgia Tech University, 
\IEEEauthorrefmark{7}University of Chicago, 
\IEEEauthorrefmark{2}North Carolina State University
}

\thanks{* The first four authors contributed equally to this work.}
\thanks{$\dagger$ The corresponding author, {zguo32@ncsu.edu}.}
\thanks{Work supported in part by NSF Grants CNS-1850851, CCF-2028481. We thank Sudharsan Vaidhun's contributions to the raw footage.}
}

\begin{document}

\maketitle
\thispagestyle{empty}
\pagestyle{empty}

%%%%%%%%%%%%%%%%%%%%%%%%%%%%%%%%%%%%%%%%%%%%%%%%%%%%%%%%%%%%%%%%%%%%%%%%%%%%%%%%
\begin{abstract}
Real-time machine learning {object} detection algorithms are often found within autonomous vehicle technology and depend on quality datasets. It is essential that these algorithms work correctly in everyday conditions as well as under strong sun glare. 
Reports indicate glare is one of the two most prominent environment-related reasons for crashes.
However, existing datasets, such as {the Laboratory for Intelligent \& Safe Automobiles Traffic Sign (LISA) Dataset} and the German Traffic Sign Recognition Benchmark, do not reflect the existence of sun glare at all. 
This paper presents the GLARE\footnote{\textbf{GLARE is available at {https://github.com/NicholasCG/GLARE\_Dataset}.}} traffic sign dataset: a collection of images with U.S-based traffic signs under heavy visual interference by sunlight. 
GLARE contains 2,157 images of traffic signs with sun glare, pulled from 33 videos of dashcam footage of roads in the United States. It provides an essential enrichment to the widely used LISA Traffic Sign dataset. 
Our experimental study shows that although several state-of-the-art baseline architectures have demonstrated {good performance on traffic sign detection in conditions without sun glare in the past, they performed poorly when tested against GLARE (e.g., average mAP$_{0.5:0.95}$ of 19.4)}. We also notice that current architectures have better detection when trained on images of traffic signs in sun glare {performance (e.g. average mAP$_{0.5:0.95}$ of 39.6),} {and perform best when trained on a mixture of conditions (e.g. average mAP$_{0.5:0.95}$ of 42.3).}

% t, the average mAP0.5 is 34.7,
% and the average mAP0.5:0.95 is 19.4.

% Our experimental study shows that although several state-of-the-art baseline methods demonstrate superior performance 
% when trained and tested against traffic sign datasets without sun glare, they greatly suffer when tested against GLARE (e.g., ranging from 9\% to 21\% mean mAP, which is significantly lower than the performances on LISA dataset).
% We also notice that current architectures
% have better detection accuracy (e.g., on average 42\% mean mAP gain for mainstream algorithms) when trained on images of traffic signs in sun glare.
\end{abstract}

\vspace{-8mm}

\section{Introduction}
\vspace{-5mm}
% Eases the reader into what the main idea will be
Driving has seen its numerous phases of evolution, from being steam-propelled to becoming almost fully autonomous. Throughout these developments, the existence of one phenomenon has remained constant in the daily environment --- intense sunlight which can obstruct the view of a vision sensor (either eyes or cameras) while maneuvering a vehicle. When the sun descends on the horizon, sun glare seeps below a car's visor and visually impairs vision sensors, causing difficulty in navigating everyday traffic. Temporary blindness (due to sun glare) causes difficulty in sensing other cars, and traffic signs, often leading to accidents.  As a result of sun glare, a recent report \cite{DoTreport} by the Department of Transportation has stated that as many as 9,000 glare-related accidents occur each year, making it one of the two most prominent environment-related reasons for crashes. The combination of harsh sun glare with common driving risk factors contributes to more crashes and congestion in day-to-day driving, leading to setbacks in implementing new automotive technologies. 

\vspace{-3mm}

% Current day technology and beginning to introduce the problem here
There has been an upsurge of autonomous vehicles driving alongside everyday drivers, such as Tesla or Google's Waymo. These self-driving vehicles make their decisions through the use of object detection algorithms, which allows the autonomous system to {locate} objects (such as traffic signs on the road) {using bounding boxes}, classify them, and make a real-time decision (machine learning~\cite{bian2022machine}) based on the algorithmic interpretation of the seen object. The functionality of these algorithms heavily depends on rich sets of data that are collected from real-world scenarios, annotated, and fed to ``teach'' algorithms what it may experience on the road. One set of data frequently used to teach algorithms within autonomous cars includes traffic signs---critical for navigating everyday traffic. While there are several datasets publicly available that focus on traffic signs in regular weather conditions, there is {\em few traffic sign dataset focusing on traffic signs with sun glare}. Our experiments indicate that when vehicles are continuously trained to recognize objects using data without sun glare, real-time algorithms within cars may fail to detect traffic signs and other objects when blinded by high-intensity visual noise, leading to catastrophe.  

% RE-WRITTEN LINES:
% Existing public traffic sign datasets have a common significant drawback: a lack of sun glare within its images. Although there are some publicly available datasets focused on traffic sign recognition, relatively few dataset of traffic signs exists that contains visible sun glare atop of it. If vehicles are continuously trained on perfect traffic sign data in which there exists no sun glare, how will real-time algorithms within these cars navigate during hours of strong glare? In the future, it is expected for autonomous vehicles to maneuver safely during all times of day, including peak hours in the morning and evening where sun glare is at its highest. However, if vehicles are continuously trained to recognize objects using nearly perfect data, there comes a point where these real-time algorithms will fail to detect traffic signs and other objects when blinded by high intensity visual noise, leading to catastrophe. 
\vspace{-3mm}

Datasets containing traffic signs with sun glare are often internal within autonomous driving companies and consequently are not publicly available for wider research purposes. While existing public datasets (such as {the Laboratory for Intelligent \& Safe Automobiles Traffic Sign (LISA) Dataset~\cite{mogelmose2012vision}}) do not contain any sun glare at all, there is an emerging need to create a public dataset with a wide variety of traffic signs with sun glare interference to fill this disparity. 
%Additionally, there does not exist current contributions to the LISA dataset, to which the GLARE dataset will build upon. 
%While filling this niche, GLARE will additionally be a publicly available set of images which can be used to train real-time detection algorithms and more. This dataset and its proposed algorithms are intended to act as a baseline for upcoming researchers while developing, training, and examining their own models.

\renewcommand{\thempfootnote}{\fnsymbol{mpfootnote}}
\begin{table*}[ht]
\begin{minipage}{\textwidth}
\caption[xxx]{Comparison of Existing Traffic Sign Datasets.}
\vspace{-3mm}
\begin{center}
\scalebox{1.1}{
\begin{tabular}{|c|c|c|c|c|c|c|c|} 
 \hline
 Dataset & Images & Image Resolution & Classes & Tasks\footnote{We target {localization} and the recognition (classification) tasks here.} & Features & Country & Year \\ [0.5ex] 
 \hline\hline
 STS~\cite{larsson2011correlating} & 3,777 & 1280$\times$960 & 20 & Both & w/ general occlusion & Sweden & 2011\\
 \hline
 GTSRB~\cite{stallkamp2012man} & 51,839 & 15$\times$15 $\sim$ 250$\times$250 & 43 & Recognition & w/ general occlusion & Germany & 2011\\
 \hline
 GTSDB~\cite{houben2013detection} & 900 & 1360$\times$1024 & 43 & {Localization} & w/ general occlusion & Germany & 2013\\
 \hline
 LISA~\cite{mogelmose2012vision} & 6,610 & 640$\times$680 $\sim$ 1024$\times$522 & 49 & Both & w/ general occlusion & USA & 2012\\
 \hline
 TT-100K~\cite{zhu2016traffic} & 100,000\footnote{Only 10,000 images contain traffic signs.} & 2048$\times$2048 & 221 & Both & w/ general occlusion & China & 2016\\
 \hline
 MTSD~\cite{ertler2020mapillary} & 100,000 & 1000$\times$1000 $\sim$ 2048$\times$2048 & 313 & Both & w/ general occlusion & Worldwide & 2019\\ 
 \hline
 DFG~\cite{tabernik2019deep}\footnote{The DFG Traffic Sign Dataset uses polygon annotations, instead of bounding box annotations.} & 6,957 & 720 $\times$576$\sim$1920$\times$1080 & 200 & Both & w/ general occlusion & Slovenia & 2019\\
 \hline \hline
 \textbf{\TheName{}} & 2,157 & 720$\times$480$\sim$ 1920$\times$1080 & 41 & Both & \textbf{w/ heavy glares} & USA & 2022\\ 
 \hline
\end{tabular}
\label{tab:comparison}}
\end{center}
\end{minipage}
\end{table*}
\vspace{-3mm}
% TO-DO MICHE: 
% CONTRIBUTIONS SHOULD BE IN BULLET POINTS - REWORK THIS
\textbf{Contributions.} As an addition to the LISA Traffic Sign dataset, we establish the GLARE dataset --- a collection of images with traffic signs which have heavy visual interference as a result of strong sunlight. GLARE will be a publicly available set of images for training real-time {object} detection algorithms and more. This dataset and the proposed algorithms are intended to act as a baseline for upcoming researchers while developing, training, and examining their own models. The contributions of this work comes in three folds:
\vspace{-6mm}
\begin{itemize}
    \item We establish a fine-grained traffic sign dataset, GLARE, abundant with realistic glares on or near the traffic sign areas. To our knowledge, GLARE is the first traffic sign dataset with detailed annotations of sun glares, covering varied scenarios of glare conditions from daily driving. Compared to the commonly used dataset (e.g., LISA~\cite{mogelmose2012vision}, GTSDB~\cite{houben2013detection}, and TT-100K~\cite{zhu2016traffic}), GLARE provides pure observation of traffic sign with glares instead mixing with a sparse witness of general occlusions. We follow the standard format to annotate, calibrate, and reorganize the dataset for a wide range of research tasks (e.g., traffic sign {localization}, image classification, and temporal localization). 
    \item We also have released the full procedures to step-by-step create the dataset and analyze its statistical features.
    \item We further showcase the research potentials of the GLARE dataset by testing it on a large group of benchmarks. Specifically, we observe that the performances of mainstream {object} detection {architectures} used in real-time traffic sign detection degrades drastically {when trained on the LISA dataset, whereas} training with the GLARE dataset shows a significant performance gain instead. %To facilitate the practical usage of w/ and w/o GLARE trained traffic sign detection algorithms, we further design a glare detector to automatically switch the using of the well-trained detection algorithms.  
\end{itemize}

% We propose that current state-of-the-art object recognition algorithms such as YOLOv5 face difficulty in recognizing everyday traffic signs found in the GLARE dataset due to the presence of strong sun glare. We use several well-known object recognition algorithms to test our theory, test and train with our dataset and the LISA dataset, which serves as a non-glare comparison. 

\textbf{Organization.} The rest of the paper is organized as follows: Section II summarizes the existing related work of traffic sign datasets and cutting-edge object detection {architectures}. Section III details the dataset including its collection, annotation, and statistics. Section IV reports the experiments to check the testing performance of the mainstream {object} detection {architectures} with {both partially and entirely} and without the GLARE dataset in the training phase. Section V concludes the paper {and suggests ideas for future research.}

%===============================================================================================================================================================================

\vspace{-5mm}
\section{Related Work}
\vspace{-2mm}
% CHECK IF 'CONNECTED COMPONENT ANALYSIS' HAS BEEN CITED - Nick, 07/19

% Jiang: these section includes the previous traffic sign datasets (focus), and the traffic sign detection algorithms (less focus)

% Related Works Papers and Their Links:
% Previous Traffic Sign Datasets
% GTSRB + GTSDB = https://ieeexplore.ieee.org/stamp/stamp.jsp?tp=&arnumber=6033395
% Mapiliary = https://arxiv.org/pdf/1909.04422.pdf 
% LISA = https://ieeexplore.ieee.org/stamp/stamp.jsp?tp=&arnumber=6335478
% RoadText1k = https://ieeexplore.ieee.org/stamp/stamp.jsp?tp=&arnumber=9196577
% CURE-TSD-Real (In TITS) = https://ieeexplore.ieee.org/stamp/stamp.jsp?tp=&arnumber=8793235
% Tsinghua-Tencent 100K = https://cg.cs.tsinghua.edu.cn/traffic-sign/ 

% Traffic Sign Detection Algorithms

\subsection{Traffic Sign Datasets}
\vspace{-3mm}
% IN PROGRESS BY MICHE; 6/27
% CHECK LISA CITATION
With the advancement of autonomous driving, there has been an emphasis on collecting data with all types of road conditions, signs, and any factor to note while driving, leading to a plethora of datasets in the community specific to traffic sign detection. 

\vspace{-4mm}

Several datasets tend to focus on traffic signs found globally, each with variations. For example, the German Traffic Sign Recognition Benchmark~\cite{houben2013detection} focuses on traffic signs from Germany and captured images in different environments under varied weather conditions. Others that follow a similar pattern include the Tsinghua-Tencent 100K dataset~\cite{zhu2016traffic}, the Swedish Traffic Sign dataset~\cite{larsson2011correlating}, and the Belgium Traffic Signs dataset~\cite{timofte2014multi}. It is advantageous to the computer vision community to have access to traffic signs from around the world, but there is a significant drawback common to public traffic sign datasets: a lack of sun glare within its images.

\vspace{-3mm}

The use of convolutional neural networks (CNNs) is prevalent throughout traffic sign datasets, often for the tasks of {localization}, recognition, {and joint localization and recognition, which is commonly referred to as object detection}. To set the standard for these tasks, baselines are often attached to datasets in the form of varied CNNs. The Mapilliary dataset~\cite{ertler2020mapillary}, for example, uses a Faster regional-based convolutional neural network (R-CNN) based detector to produce mean average precision (mAP) results over all of its classes. The DFG Traffic Sign dataset~\cite{tabernik2019deep} is another example that uses such techniques to establish a baseline, utilizing a Faster R-CNN and a Mask R-CNN to provide mAP values ranging in the upper 90s.
%results. Often, the resulting mAP values range in the upper 90's when tested upon images of traffic signs within the DFG Traffic Sign dataset. 
Although the dataset includes traffic-sign instances with synthetic distortions that may resemble sun glare, these images are incomparable to those with natural sun glare. Generalization performance may suffer from improper training when strictly utilizing datasets that lack natural sun glare. This is a phenomenon found often throughout datasets focused on traffic sign depiction:  models are trained on data without obscuring conditions or with synthetic ones, such as sun glare, usually %provide exceptional results which usually 
are unsatisfactory when tested in real driving scenarios. 

\vspace{-3mm}

%As we are concerned, s
Severe conditions, such as sun glare or heavy rain, impede the visibility of traffic signs while driving. Just as it hinders human drivers, it additionally interferes with algorithmic vision. The CURE-TSD-Real dataset~\cite{temel2019traffic}, comprised of traffic sign images in simulated heavy road conditions, is an example of a dataset with severe conditions that resulted in a 29\% drop in average precision. 
%This kind of downgrade is seen with images of traffic signs captured under conditions such as haze, snow, or rain. Thus, we are 
This motivates us to investigate the possibility of harsh sun glare causing a drop in algorithmic performance as well.
%similarly to how other visibility impairing conditions have done. 

\vspace{-3mm}

The GLARE dataset intends to be an extension of the LISA dataset, which is one of the most commonly used American traffic sign datasets with an emphasis on large variations within urban landscapes. The dataset is comprised of videos and stand-alone images of traffic signs, amounting to about 6,610 images and 7,855 annotations. Source data for the LISA dataset comes with color, in grayscale, and does not include images with excessive sun glare. The LISA dataset answered a need for a public dataset with US-based traffic signs and notably contributed more as it includes full traceability of its dataset by providing full annotations of all images, and includes all associated tracks. We provide full comparisons of the existing related traffic sign datasets in Table~\ref{tab:comparison} with GLARE, which shows that GLARE is the latest traffic dataset (with two years gap from DFG and MTSD and nine years gap from original LISA) and is formed by high-resolution images with heavy/harsh glares on traffic signs.

% Datasets are often constructed for usage for object detection or recognition. Often times, there is a distinction between the two... 

% %Miche things to write about"
% Conditions under which these datasetss use are..... but none include glare specifically
% Road text is one that has fog? and it brings down its values so there is validity in showing how different weather conditions can cause change in image processing...
% [Discuss what details do these annotations have and or how other datasets dont have occulence in their stuff -refer to mapiliary for more]
\vspace{-6mm}
\subsection{Traffic Sign Classification}
\vspace{-4mm}

One of the most popular applications with the aforementioned datasets is traffic sign classification, where tremendous efforts are accomplished from statistical learning to deep learning paradigms. For example, Soendoro and Supriana~\cite{soendoro2011traffic} first adopt the SVMs with sparse representation to recognize the class of traffic signs in images. With the rise the deep learning, convolutional neural networks begin to dominate the performances of recognition/classification in the traffic sign domain. Specifically, on the GTSRB dataset, a large amount of CNN variants~\cite{mao2016hierarchical} shows a powerful ability for generalization, where the classification accuracy on the testing set is even better than the performance of human experts (e.g., CNNs with spatial transformers~\cite{arcos2018deep} can achieve roughly 99.7\% in terms of top-1 accuracy). Note that, the reported high classification accuracy is based on the cropped traffic sign images, where we can obtain these images via a specifically designed {object} detection task.

\vspace{-6mm}
\subsection{End-to-End Traffic Sign Detection}
\vspace{-4mm}

At first, the {localization} and the classification are two independent tasks, where the classification is built upon the properly {localized bounding boxes} (i.e., the traffic sign is located and extracted intact).
With the rise of the CNNs, the original powerful performance of image classification/recognition~\cite{krizhevsky2012imagenet} rapidly transfers to object detection domain. Furthermore, it has been a consensus that the family of CNNs is capable of detecting a bounding box for a specific object while classifying its category simultaneously. The well-known RCNN/Fast-RCNN~\cite{girshick2014rich,girshick2015fast} first generates potential bounding boxes on the frames and then classifies the object only in these bounding boxes. However, the final performance of {object} detection depends on the performances of multiple stages during the complex pipeline (i.e., pre-processing, classification, and post-processing to re-score the proposed bounding boxes), where the whole process is slow as well. To address the efficiency and complexity issue, ~\cite{redmon2016you} proposes the first edition of You Only Look Once (YOLO) series, v1 to v5, and treats the object detection as a single regression task to directly establish the connection among the image pixels, the bounding box coordinates and the labels with probabilities. To further improve the performance, YOLOX~\cite{ge2021yolox} is proposed as incorporating the anchor-free manner and several cutting-edge detection techniques (e.g., decoupled head, {dynamic} label assignment strategy).  

\vspace{-3mm}

Another branch of {object} detection strategies leverages the popular transformer~\cite{carion2020end} encoder-decoder architectures by removing the complicated hand-designed components such as non-maximum suppression or anchor generation while optimizing a global loss  that enables \color{black} unique classifications via bipartite matching. Similar ideas are brought into traditional RCNN architecture that a special Swin Transformer~\cite{liu2021swin} shows a great performance gain when replacing the ResNet50 backbone. Almost all the aforementioned {object} detection algorithms rely on supervised learning with labeled traffic signs and it is rarely considered that the possible strong {localized} noises (e.g., sun glare) in the testing phase may degrade the detection performance.

\vspace{-6mm}

\section{GLARE Dataset}
\vspace{-4mm}

This section presents the GLARE traffic sign dataset, a sun glare focused dataset to assist researchers and developers in building real-time autonomous traffic sign detection and classification system in sun glare conditions. The dataset includes 2157 images and annotations, each containing a single traffic sign annotation. This dataset can be used for {object localization, recognition, and object detection tasks}. 
%Here we will describe how we collected and processed the videos and images, handle the annotation process, and report the statistics about the GLARE dataset.

% Kurt's camera: Rexing V1P
% Jiang's camera: idk
\vspace{-6mm}
\subsection{Video Collection and Processing}
\vspace{-4mm}
The initial collection process started with three dashboard cameras recording approximately 38 hours of footage around the Orlando area. Two cameras were forward facing with one filmed at 1920$\times$1080 (1080p) and the other rearward facing one filmed at 720$\times$480 (480p). In total, the cameras filmed 463 initial videos of 40 hours and 25 minutes of footage.
%40:25:39 in precise total.

\vspace{-3mm}

The first step in video processing was to remove all videos that did not meet the criteria of having both sun glare and traffic signs at the time. %Therefore, all videos that either had no sun glare present or were in environments where sun glare could not be present were discarded. After this initial processing step, there were 
This resulted in 163 videos that contained some amount of sun glare. The second step was to extract the sections of video with sun glare, referred to as {\em clips}. The clips each contained a continuous presence of sun glare, with about a half second of extra time at the start of each clip to allow for ease of finding the beginning of the sun glare. 
%After extracting the clips from the original footage with sun glare, further footage was extracted that 
Since only footage that concurrently contains sun glare and traffic signs matters, any clips that did not contain traffic signs during the follow-up screening section were discarded. Following a similar procedure as the LISA dataset~\cite{mogelmose2012vision}, these short videos are referred to as {\em tracks}. 189 tracks were used in the creation of the GLARE dataset, totaling 18 minutes and 11 seconds of footage. % from the original footage.
The tracks were organized by their original source video, with 33 original source videos being used in total to produce the GLARE dataset. 
%In the following subsection, we will introduce the process of annotation on these collected tracks of videos.

\vspace{-6mm}
\subsection{Annotation Process}
\vspace{-4mm}
The image annotation process was separated into two main steps: bounding box localization of traffic signs, and bounding box approval with cleanup. The first step was completed by two individuals at the same time with a single open-source tool~\cite{videolabel} to allow for efficient labeling. The second step was completed after the initial processing of all the images as a quality assurance step by two individuals who worked on labeling and processing the initial bounding boxes in the LISA dataset format~\cite{mogelmose2012vision}. 
%The first two steps were completed by two individuals and the third step was completed by one of the individuals who worked on labeling and processing the initial bounding boxes.
%Both the bounding box localization and the bounding box approval process were completed using a custom video player and annotator developed from an existing open-source program~\cite{videolabel}. The annotator allowed for bounding box annotation, automatic bounding box tracking with stepped saving of bounding boxes, bounding box classification, bounding box approval and deletion, and the exportation of bounding boxes and associated annotation data in the LISA dataset format~\cite{mogelmose2012vision}. % (see Fig.~\ref{fig:opencv-vid-label}). 
The automatic bounding box tracking algorithms available with the tool were Re3~\cite{gordon2018re3} and CMT (Consensus-based Matching and Tracking of Keypoints for Object Tracking)~\cite{Nebehay2015CVPR}, and we used Re3 for all our annotations due to stable tracking at all sizes. When annotations were saved, each image that was exported had a single associated annotation. %each annotation would be associated with a single separate image, meaning
\vspace{-3mm}

\noindent{\bf Traffic Sign Localization.} In the first step, tracks were processed together based on the original source video. 
Each track would then be played to completion, with the bounding box labeling occurring on the first frame with sun glare that was a multiple of 5. The process would continue with the current bounding box being saved on every subsequent multiple of 5 until the track finished playing or sun glare was no longer on the screen.
%For each track, the track would be played to completion, with the bounding box labeling beginning on the first frame with sun glare that was a multiple of 5, and with every current bounding box being saved on each multiple of 5 frame until either the track finished playing or the sun glare was no longer on screen. 
When labeling traffic signs using the bounding box localization tool, you could automatically choose the classification of the traffic sign to allow for increased efficiency. %(see Fig.~\ref{fig:processer}). 
After the initial labeling, the annotation tool would continue automatically annotating until the current user deemed the automatic annotation to have drifted too far. The annotator would then delete and reapply the bounding box, and continue annotating until all the tracks were processed. For each traffic sign in a track, no more than 30 annotations of that traffic sign would be saved to decrease the overexposure of that sign.

\begin{figure}[ht]
% \begin{subfigure}{.49\textwidth}
%   \centering
%   % include first image
%   \includegraphics[width=.9\linewidth]{Images/video_labeler.png}  
%   \caption{Bounding box selection, classification, and automatic tracking.}
%   \label{fig:labeller}
% \end{subfigure}
%\begin{subfigure}{.5\textwidth}
  \centering
  \vspace{-3mm}
  % include second image
  \includegraphics[width=.99\linewidth]{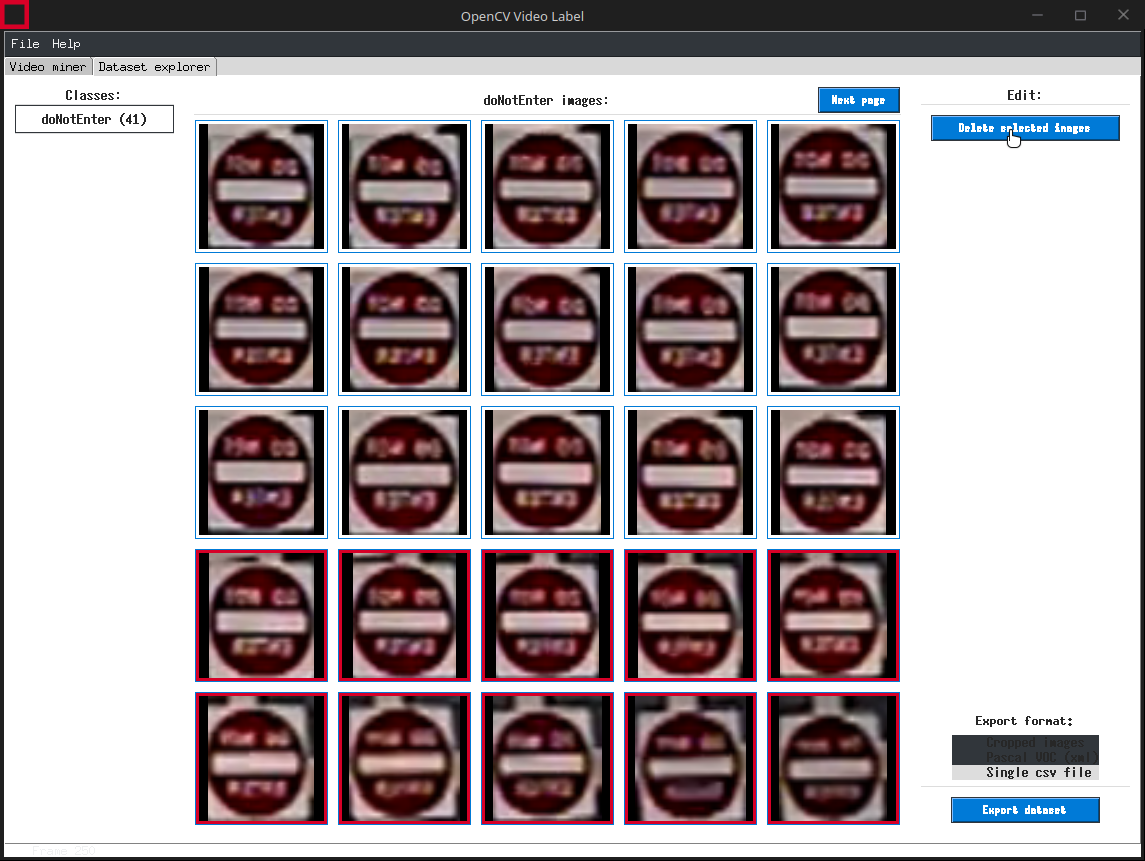}  
  \caption{Bounding box processing and exporting.}
  \vspace{-2mm}
  \label{fig:processer}
%\end{subfigure}
%\caption{Example of our custom version of OpenCV Video Label~\cite{videolabel} for easier classification and bounding box processing.}
%\label{fig:opencv-vid-label}
\end{figure}

\noindent{\bf Bounding Box Approval and Cleanup.}
After a track was labeled, the annotations were reviewed and either approved or rejected. Any bounding boxes with background noise that can be removed with manual selection are removed and relabeled. After all tracks and annotations were processed, %and annotations approved or rejected, 
the video was exported in a single CSV file (similar to the LISA dataset) for further processing, as demonstrated by Figure~\ref{fig:processer}. %for an example. The exported annotations include the same information included in the LISA dataset in a single CSV file, except the "Occluded" and "On another road" annotations~\cite{mogelmose2012vision}, which are added in the following step.

%\subsubsection{Annotation Processing and Cleanup}

After all the tracks were processed and exported by the original source video, the annotations were further processed to remove previously uncaught errors %, fix any errors in classification, 
and extract statistical information from the entire dataset. 
%Each annotation was checked for having the proper classification and having a well bounded bounding box. 
Any bounding box that did not localize a sign or contained significant background noise was rejected, and any improperly labeled annotations were renamed. After removing the improper annotations, the remaining annotations were then categorized on if the traffic signs were covered in any way, and if they were on the current road or a side road~\cite{mogelmose2012vision}. These "Occluded" and "On another road" annotations~\cite{mogelmose2012vision} were then pooled.
%, followed by extracting and important data statistics were extracted, which will be discussed in the following section.
\vspace{-7mm}
\subsection{Dataset Statistics}
\begin{figure}[ht] 
\centering
\vspace{-5mm}% <---
   \begin{subfigure}{0.49\textwidth}
       \centering
        \includegraphics[width=\linewidth]{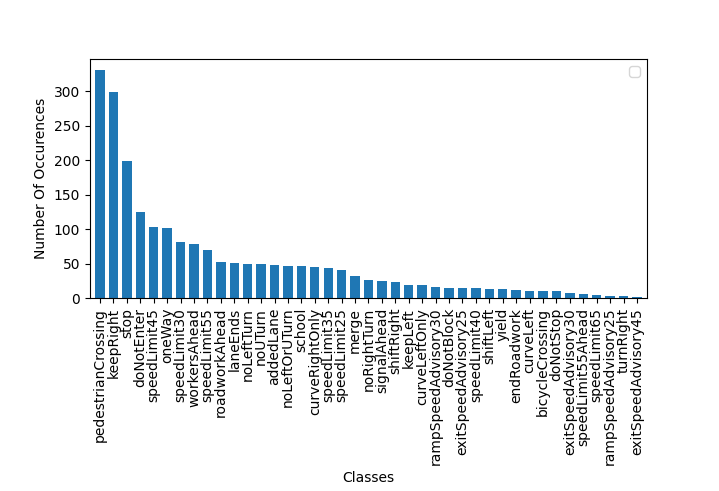}
        \vspace{-6mm}
        \caption{Traffic sign classes.}
        \label{fig:counts}
   \end{subfigure} \\ \vspace{-0.5mm}
%\hfill % <---  
   \begin{subfigure}{0.49\textwidth}
       \centering
        \includegraphics[width=\linewidth]{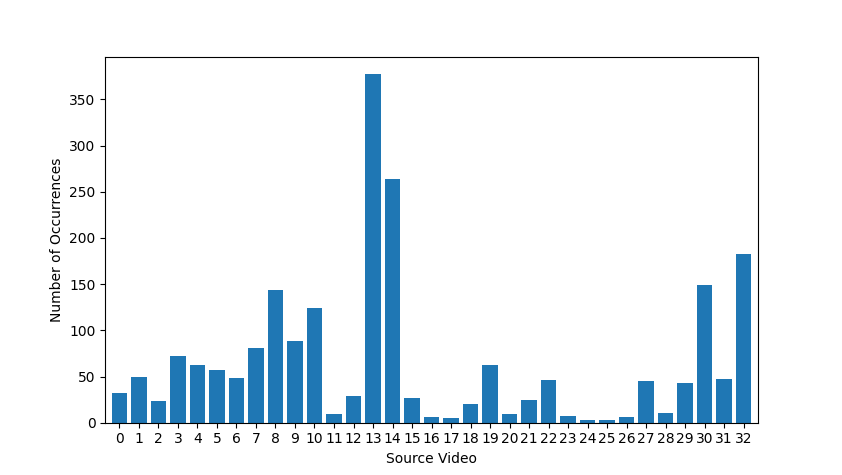}
        \vspace{-6mm}
        \caption{Annotations per source video.}
        \vspace{-1mm}
        \label{fig:annot_counts}
   \end{subfigure}
   \caption{Statistics/Distributions of GLARE dataset.}
   \label{fig:image2}
   \vspace{-4mm}
\end{figure}

\vspace{-4mm}
The GLARE dataset contains 2,157 bounding box annotations and associated images distributed across 41 classes. Figure~\ref{fig:counts} shows the distributions of the annotations per class. The annotations were created from multiple videos to ensure a variety in the location in glare conditions. For each track in each source video, a maximum of 30 frames for each traffic sign class were allowed to minimize over-exposure of traffic signs in specific positions and sun glare conditions. Figure~\ref{fig:annot_counts} shows the distribution of annotations across the 33 source videos processed. The size of the bounding box annotations varies between 6 $\times$ 14 and 137 $\times$ 178 pixels, and the size of the images is either 810 $\times$ 540 or 937 $\times$ 540 pixels. %Due to the relatively small number of annotations, the GLARE dataset has not been released split into training, testing, and validation sets. However, 
The dataset works with existing scripts released alongside the LISA dataset for annotation, extraction, and splitting~\cite{mogelmose2012vision}.

The types of visual interference labeled as sun glare can be broadly categorized into four categories. The categories described are subjective but are recorded to allow for a greater understanding of how we evaluated sun glare during the initial video processing for the dataset. Examples of each category can be seen in Figure~\ref{fig:image2}. The first category is where there is a clear sun without any significant additional bright cloud noise or brightness interference from the camera. The sun appears as a bright ball, excluding any obstruction by either clouds or other objects. In an upcoming section, we will describe a naïve detector for this type of sun glare to improve traffic sign detection results. The second category is where there is a visible sun, but there are additional clouds that add to the overall brightness of the image. The third category is where there is minimal to no visible sun due to cloud interference. Although the sun is not visible in the frame, there is still visual interference that causes traffic signs to be less visible than in clear conditions, decreasing detection. The fourth category is sun glare due to other interference. The sun being visible is not a requirement, as there is visual interference due to the camera settings. Either way, the visual interference appears similar to the interference caused by the other types of sun glare. The images themselves have not been labeled based on the type of sun glare due to the subjective nature of the categories and that some images can fit into multiple categories.

\begin{figure}[ht] % <---
   \begin{subfigure}{0.22\textwidth}
       \includegraphics[width=\linewidth]{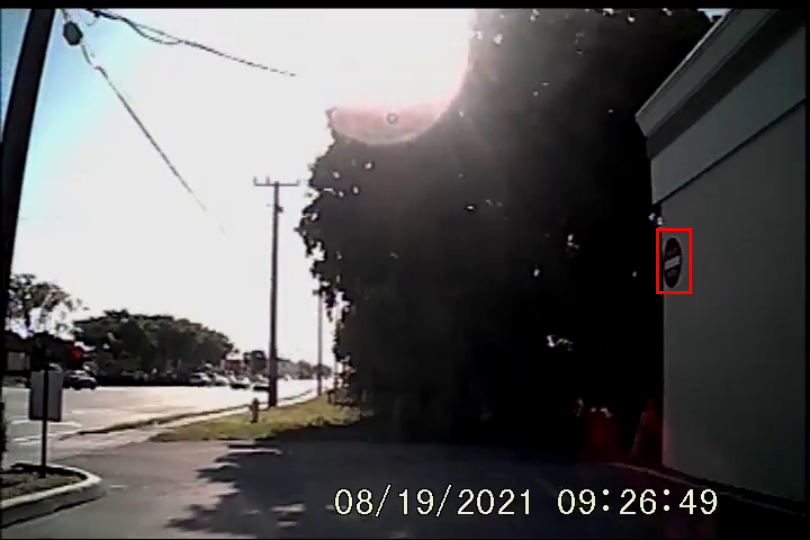}
       \caption{Sun with other interference}
   \end{subfigure}
\hfill % <--- 
   \begin{subfigure}{0.24\textwidth}
       \includegraphics[width=\linewidth]{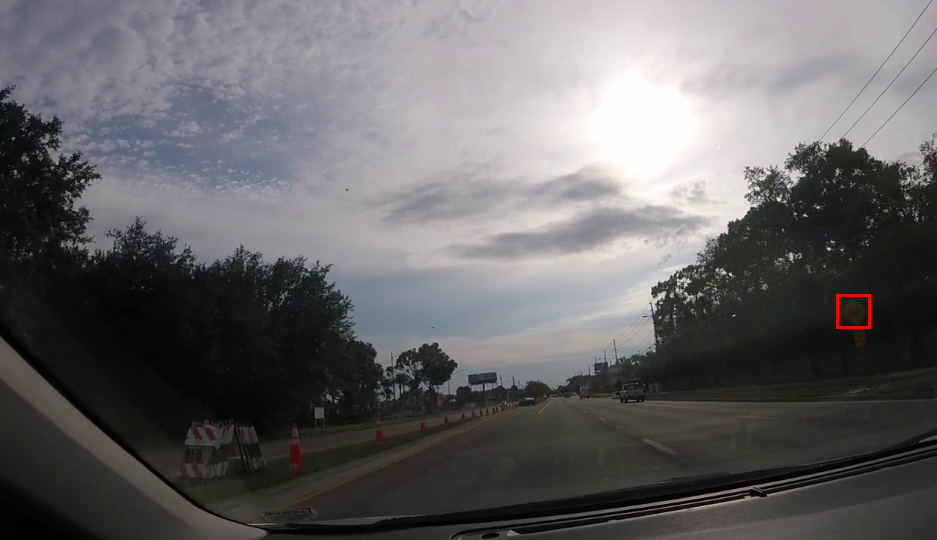}
       \caption{Clear sun without clouds}
   \end{subfigure}
\hfill % <---
   \begin{subfigure}{0.24\textwidth}
       \includegraphics[width=\linewidth]{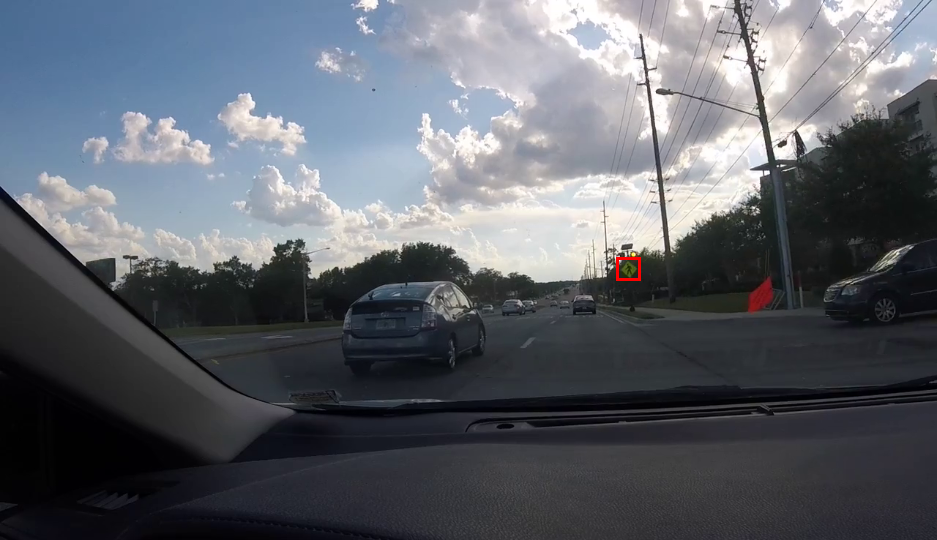}
       \caption{Clouds with non-visible sun}
   \end{subfigure}
\hfill %
   \begin{subfigure}{0.24\textwidth}
       \includegraphics[width=\linewidth]{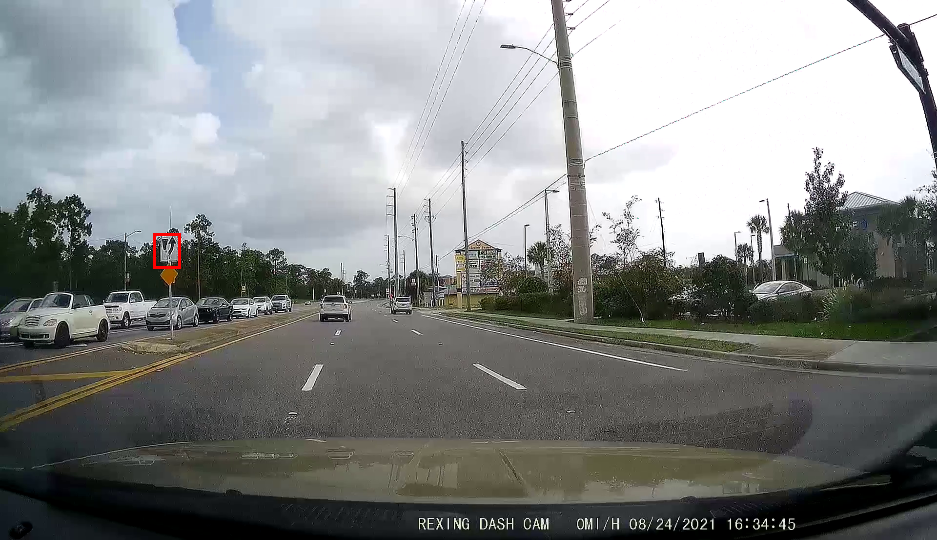}
       \caption{Clear sun with clouds}
   \end{subfigure}
   
   \caption{Examples of images form the GLARE dataset with bounding boxes highlighted.}
   \label{fig:image2}
\end{figure}

\vspace{-4mm}
\section{Benchmarks}
\vspace{-4mm}

In order to test how sun glare conditions affect the ability to detect traffic signs, we performed multiple tests comparing how different state-of-the-art object detection architectures perform detecting traffic signs in sun glare conditions. Each architecture was tested by being trained on only the LISA dataset, on only the GLARE dataset, and on the LISA and GLARE datasets combined. The trained models were tested on similar testing sets, with the LISA-trained models tested on a subset of the testing set for the GLARE and combined models. We also performed a supplementary test using the YOLOv8 architecture for comparing how using initially random weights and using pre-trained weights affects the ability to detect traffic signs in sun glare conditions.
\vspace{-6mm}
\subsection{Splitting Methodology}
\vspace{-4mm}
To fairly test all the trained models and minimize similar frames in the training sets and testing set, the initial testing set was created using the frames from 5 videos in the GLARE dataset with a focus on maximizing traffic sign coverage for sign types present in multiple videos while minimizing the percentage of the GLARE dataset used. However, it was found that about half of the traffic sign types are only present in a single video. Therefore, to give coverage of these traffic sign types in the testing set while minimizing the likelihood of similar frames across the sets, the last frames for traffic sign types with less than 5 frames, in the order they appear in the original video, were selected for the testing set. This resulted in about 26.52\% of the GLARE dataset being used for the testing set.

To train each architecture, 3 different training sets were created for the GLARE-only models, the LISA-only models, and the combined dataset models. For the LISA dataset models, the training set was created by selecting a subset of the LISA dataset with only the frames with traffic sign types that are present in both the LISA dataset and the GLARE dataset. Similarly, a custom testing set was also created for these models by subsetting the larger testing set with only the frames containing traffic sign types present in both GLARE and LISA. This was to not test the models on traffic sign types not present during training. For the GLARE dataset models, 3 augmented copies for each frame in GLARE not selected for testing were created with alterations using noise, color jitter, and blur, and added to the training set along with the original image. The GLARE training set was then split into 3 folds to account for the added bias of similar frames in the training set with 3 fold cross-validation performed. The resulting models were tested using the entire testing set with the scoring results averaged. For the combined models, the training sets for the LISA and GLARE models were combined, shuffled, and split into 3 folds similar to the GLARE training set for 3 fold cross-validation. Similarly, the resulting models were tested using the entire testing set with the scoring results averaged.
%Therefore, there are 7855 images in the training set, and 1613 images in the validation set for these models. For the models trained on the GLARE dataset, the training and validation sets are split on an 80/20 ratio, with each class of traffic sign being in each set. Therefore, there are 1725 images in the training set and 432 images in the validation set for these models.  

% \usepackage{booktabs}

\color{black}
% This feels rough, a little guidance in explaining mAP would be helpful -Nick
\vspace{-6mm}
\subsection{Scoring Methodology}
\vspace{-4mm}
% The scoring metric used for comparing the performance of the models is the mean average precision (mAP) of the classes as defined in COCO~\cite{lin2014microsoft} and implemented in Ultralytics' YOLOv5~\cite{glenn_jocher_2022_6222936} and OpenMMLab~\cite{mmdetection}. For each image, we produce multiple predicted bounding boxes, and compare the intersection of each bounding box to the ground truth of the traffic sign in the image. If the predicted bounding box's Intersection over Union (IoU) with the ground truth is over 0.5, then the prediction is counted as positive. In the case of multiple prediction to a ground truth bounding box, only the prediction with the highest IoU is counted. 
% %so as to not have multiple predictions to one ground truth. After collecting the predicted bounding boxes and sorting them by either being true or false positives (TP and FP), we sort the predictions by their confidence score along with sorting the TPs and FPs in descending order. After sorting, the precision and recall values are calculated using the TPs and FPs. 
% The Average Precision (AP) is then calculated as the area under the precision-recall curve,
% %To calculate for the mAP, each class is calculated separately with their own AP score in a binary fashion, 
% %and the APs are meaned together to produce the mAP over all classes. We
% and over all classes, we calculate mAP$_{0.5}$ as the mean over all classes with an IoU of 0.5 and mAP$_{0.5:0.95}$ as the mean over all the classes over a range between 0.5 to 0.95, with a step of 0.05~\cite{lin2014microsoft}.

The scoring metric used for comparing the performance of the models is the Mean Average Precision (mAP) of the classes as defined in COCO~\cite{lin2014microsoft} and implemented in Ultralytics' YOLOv5~\cite{glenn_jocher_2022_6222936} { and YOLOv8~\cite{Jocher_YOLO_by_Ultralytics_2023} releases} and OpenMMLab’s mmdetection toolbox~\cite{mmdetection}. For each {ground truth bounding box in an} image, multiple predicted bounding boxes {are produced, and the ratio between the area of the intersection between each prediction and the ground truth and the area of the union between each prediction and the ground truth is calculated.} If {a} predicted bounding box’s Intersection over Union (IoU) is equal to over {a specified threshold}, then the prediction is {kept with the IoU used as a confidence score.} The Average Precision (AP) is then calculated as the area under the precision-recall curve {for the kept predictions, and the Mean Average Precision is the average of the Average Precision over all the different label types. We calculate mAP$_{0.5}$ as the Mean Average Precision where the specified threshold for the IoU is 0.5. mAP$_{0.5:0.95}$ is calculated as the average of the Mean Average Precision with the specified thresholds being over a range from 0.5 to 0.95 inclusive,} with a step of 0.05~\cite{lin2014microsoft}.

\vspace{-6mm}
\subsection{Configuration and Implementation}
\vspace{-4mm}

To demonstrate the necessity of viability of the GLARE dataset, we experimented with seven state-of-the-art methods comparing the results when training on only GLARE, only LISA, and GLARE and LISA together. We chose to test on a variety of different architecture families, with a focus on including state-of-the-art one-stage architectures commonly used in real-time object detection tasks and architectures based on the transformer architecture, which has demonstrated greatly improved results on several deep learning tasks, such as object detection. For this, we chose YOLOv5~\cite{glenn_jocher_2022_6222936}, YOLOX~\cite{ge2021yolox}, YOLOv8~\cite{Jocher_YOLO_by_Ultralytics_2023}, and TOOD~\cite{feng2021tood} as our one-stage detection architectures, and chose Deformable DETR~\cite{zhu2021deformable} and the Faster-RCNN~\cite{Ren_2017} with a Swin Transformer backbone~\cite{liu2021swin} as our transformer architectures. Finally, we also selected the Faster-RCNN with a ResNet50 backbone~\cite{resnet} as a baseline to compare the more recent and state-of-the-art architectures against.

The three YOLO architectures were chosen due to YOLOv8 and YOLOX having architecture changes to improve results compared to YOLOv5. For example, YOLOv8 and YOLOX are anchor-free architectures while YOLOv5 is anchor based, which tends to increase the number of bounding box predictions. YOLOv8 has also done other changes compared to YOLOv5, such as changing the first convolution layer in the stem from 6x6 to 3x3 kernel, reducing the CSP Bottleneck from 3 convolutions to 2, and changing the kernel size of the first convolution layer in the bottleneck to 3x3. YOLOX also differs from YOLOv5 by utilizing a decoupled head and a novel dynamic label assignment strategy named simOTA. All 3 YOLO architectures were trained using the small model size, as the GLARE dataset is designed to be used in real-time object detection. TOOD was chosen to demonstrate the performance of a one-stage detection architecture that performs object detection differently from the YOLO series, as TOOD uses a “new head structure and alignment-based learning approach”~\cite{feng2021tood} to align the classification and localization tasks for object detection.

The YOLOv5 and YOLOv8 models were trained and tested on Ultralytics' releases~\cite{Jocher_YOLOv5_by_Ultralytics_2020,Jocher_YOLO_by_Ultralytics_2023}, and the rest of the architectures were trained and tested using OpenMMLab's MMDetection toolbox~\cite{mmdetection}. All architectures used the given training and testing pipelines for images to not bias the detection results to any training set. As MMDetection provides different configurations for the available architectures, TOOD was trained with multi-scale training, Deformable DETR used the 2-stage version with iterative box refinement, and the Faster-RCNN with a Swin Transformer backbone used multi-scale cropping. 

The Faster-RCNN, Deformable DETR, and TOOD architectures used the ResNet50 backbone with pre-trained weights trained on the COCO dataset~\cite{lin2014microsoft}, and the Faster-RCNN with a Swin Transformer backbone used pre-trained weights trained from the ImageNet-1k dataset~\cite{deng2009imagenet} for the backbone. All the YOLO architectures were trained with completely random weights in the backbone. To account for the possible decrease in performance compared to the other architectures with pre-trained weights, we performed a supplementary test comparing the performance of YOLOv8 when trained using random weights and using weights pre-trained on the COCO dataset, which the results are described in the following section.

The YOLOv5 models were trained on an RTX 3070, the YOLOv8 and TOOD models trained on LISA and GLARE were trained on an RTX 3090, and the rest of the models were trained on two NVIDIA Tesla V100 GPUs. For all the architectures, we used the default hyperparameters provided, with alterations to the training batch size, initial learning rate, and the number of epochs to fit our datasets. We only altered the learning schedule for the Faster-RCNN with a ResNet50 backbone, where we did not have any warm-up epochs. For the architectures trained using MMDetection, the auto-scale-learning-rate flag was used to automatically adjust the learning rate to the batch size. The architecture training configurations are shown in Table~\ref{table:configs}.

\begin{table}[ht]
\centering
\caption{Training configuration for each architecture}
\scalebox{1}{
\begin{tabular}{@{}lccc@{}}
\toprule
                                    & \multicolumn{1}{c}{Batch Size} & \multicolumn{1}{c}{LR} & \multicolumn{1}{c}{Epochs} \\\midrule
Faster-RCNN$_{\textit{ResNet50}}$   & 16                             & 0.02                   & 24                         \\ \midrule

YOLOv5                              & 32                             & 0.01                   & 150                        \\ \midrule

Deformable DETR                     & 8                              & 5e-5                   & 50                         \\ \midrule

Faster-RCNN$_{\textit{SwinT-Base}}$ & 16                             & 1e-4                   & 36                         \\ \midrule

YOLOX                               & 32                             & 5e-3                   & 150                        \\ \midrule

TOOD                                & 8                              & 5e-3                   & 24                         \\ \midrule

YOLOv8                              & 32                             & 0.01                   & 150                        \\ \bottomrule
\end{tabular}}
\label{table:configs}
\end{table}

\vspace{-7mm}

\subsection{Benchmark Results} 
\vspace{-4mm}
% TODO: fix The results of testing the GLARE dataset against baseline models trained on the dataset versus trained on the LISA dataset are 

\begin{table}[ht]
\centering
\caption{mAP$_{0.5}$ scoring results after training with random initial weights}
\scalebox{1}{
\begin{tabular}{@{}lccc@{}}
\toprule
                             & GLARE  & LISA    & Combined \\ \midrule
                             
Faster-RCNN$_{\textit{ResNet50}}$         & 55.1  	 & 35.3     & 68.3      \\ \midrule

YOLOv5                       & 54.8      & 26.9   & 62.6   \\ \midrule

Deformable DETR              & 62.0     & 42.1      & 71.6    \\ \midrule

Faster-RCNN$_{\textit{SwinT-Base}}$ & \textbf{68.7}  & 39.3      & \textbf{73.9}      \\ \midrule

YOLOX                        & 51.5     & 17.0    & 60.7   \\ \midrule

TOOD                         & 63.5     & \textbf{57.4}      & 73.6   \\ \midrule

YOLOv8                       & 58.7     & 24.8    & 64.6   \\ \bottomrule

\end{tabular}}
\label{table:benchmarks_50}
\end{table}

\begin{table}[ht]
\centering
\caption{mAP$_{0.5:0.95}$ scoring results after training with random initial weights}
\scalebox{1}{
\begin{tabular}{@{}lccc@{}}
\toprule
                             & GLARE  & LISA    & Combined \\ \midrule
Faster-RCNN$_{\textit{ResNet50}}$            & 35.8  	 & 19.7     & 44.1      \\ \midrule

YOLOv5                                       & 37.5   & 14.9   & 42.4  \\ \midrule

Deformable DETR                              & 40.8   & 24.1   & 46.8  \\ \midrule

Faster-RCNN$_{\textit{SwinT-Base}}$          & \textbf{45.2}  & 20.6    & 48.7     \\ \midrule

YOLOX                                        & 34.6  & 10.0  & 41.1  \\ \midrule

TOOD                                         & 42.6     & \textbf{31.2}   & \textbf{49.8}   \\ \midrule

YOLOv8                                       & 41.0   & 15.0  & 43.9   \\ \bottomrule
\end{tabular}}
\label{table:benchmarks_5095}
%\vspace{-4mm}
\end{table}

\begin{table}[ht]
\centering
\caption{Scoring results after training YOLOv8 with random weights and pre-trained weights}
\scalebox{1}{
\begin{tabular}{@{}lcccc@{}}
\toprule
                             & \multicolumn{2}{c}{mAP$_{0.5}$}                   & \multicolumn{2}{c}{mAP$_{0.5:0.95}$} \\\midrule
                             & Random  &  \multicolumn{1}{l|}{COCO} & Random      & COCO  \\ \midrule
GLARE           & 58.7 	    & \multicolumn{1}{r|}{\textbf{61.7}}         & 41.0      & \textbf{44.4}     \\ \midrule
LISA            & \textbf{24.7}     & \multicolumn{1}{r|}{22.3}        & \textbf{15.0}         & 12.6    \\ \midrule
Combined        & 64.6    & \multicolumn{1}{r|}{\textbf{66.0}}        & 43.9         & \textbf{45.8}   \\ \bottomrule
\end{tabular}}
\label{table:benchmarks_coco}
\end{table}

% i will reorder for based on release
The results of testing the benchmark architectures trained on the GLARE, LISA, and combined training sets are shown in Tables~\ref{table:benchmarks_50} and ~\ref{table:benchmarks_5095}.\footnote{The architectures are listed by the initial release of the associated publication or code itself if no publication is available.}. The results of testing the YOLOv8 architecture on random weights and pre-trained weights from training on the COCO dataset are shown in Table~\ref{table:benchmarks_coco}.

For the models trained on the GLARE dataset, the average mAP$_{0.5}$ is 59.2, and the average mAP$_{0.5:0.95}$ is 39.6. For the models trained on the LISA dataset, the average mAP$_{0.5}$ is 34.7, and the average mAP$_{0.5:0.95}$ is 19.4. For the models trained on the combined GLARE and LISA datasets, the average mAP$_{0.5}$ is 67.9, and the average mAP$_{0.5:0.95}$ is 42.3. The difference between the GLARE-trained models and the LISA-trained models is 24.5 for mAP$_{0.5}$ and 20.2 for mAP$_{0.5:0.95}$, and the difference between combined-dataset-trained models and GLARE-trained models is 8.7 for mAP$_{0.5}$ and 2.7 for mAP$_{0.5:0.95}$. 

The best-performing architectures overall were the Faster-RCNN with a Swin Transformer backbone and TOOD, with the Swin architecture performing best when trained on only GLARE for both mAP$_{0.5}$ and mAP$_{0.5:0.95}$, TOOD performed best when trained on only LISA for both mAP$_{0.5}$ and mAP$_{0.5:0.95}$, and among the models trained on the combined dataset, the Swin architecture performed best on mAP$_{0.5}$ and TOOD performed best on mAP$_{0.5:0.95}$. Among transformer-based models, the Swin architecture performed best when trained on only GLARE or the combined dataset, and Deformable DETR performed best when trained on only the LISA dataset. For single-stage architectures, TOOD outperformed every architecture in the YOLO family for GLARE-trained, LISA-trained, and combined-dataset-trained models. Compared to the Faster-RCNN with a ResNet50 backbone, YOLOv5 and YOLOX performed worse when trained on only the GLARE dataset, and all architectures in the YOLO family performed worse when trained on only the LISA dataset and when trained on the combined dataset.

These results indicate that sun glare has a noticeable effect on the ability of object detection architecture to detect traffic signs in sun glare conditions, especially for the YOLO family of architectures, which are frequently used in real-time object detection tasks. When the architectures were trained on LISA alone, there was a significant decrease in performance compared to when they are trained on GLARE alone or GLARE and LISA together. Our results also validate the GLARE dataset as a useful extension of the LISA dataset, as training using the LISA and GLARE datasets together resulted in superior performance compared to both the architectures trained only on GLARE and the architectures trained only on LISA. The unusually high performance of the TOOD architecture, especially when trained on only the LISA dataset, warrants future investigation on the ability of architectures with similar structures to detect traffic signs in sun glare conditions when trained on datasets without a significant presence of sun glare.

For the test using YOLOv8 comparing using random weights and COCO pre-trained weights, using pre-trained weights led to an increase in performance by 3 in mAP$_{0.5}$ and 3.4 in mAP$_{0.5:0.95}$ for the models trained on the GLARE dataset and by 1.4 in mAP$_{0.5}$ and 1.9 in mAP$_{0.5}$ for the models trained on the combined dataset. For the models trained on the LISA dataset, training using pre-trained weights led to a decrease in performance by 2.4 in mAP$_{0.5}$ and mAP$_{0.5:0.95}$. A reasonable explanation for these results is that the pre-trained weights allow for YOLOv8, and by extension possibly other architectures in the YOLO family and similar architectures, to better fit the training set. This would lead to the models trained on LISA having more difficulty detecting traffic signs in sun glare conditions, while models trained with at least part of the data containing traffic signs in sun glare lead to increased performance, further indicating the usefulness of the GLARE dataset in traffic sign detection.

\color{black}

\vspace{-5mm}
\section{Conclusion and Future Works}
\vspace{-4mm}
This paper introduces GLARE, a traffic sign dataset with a focus on sun glare and how it affects the recognition of traffic signs in such conditions. The dataset includes 2,157 images with corresponding bounding box annotations of traffic signs across 41 classes from the Orlando area. The GLARE dataset has a specific focus on images with sun glare present, which affects both human drivers and cameras for autonomous driving systems. Our baseline benchmarks have shown that sun glare has a noticeable effect on the ability of current architectures to detect traffic signs.%, but with newer architectures having better detection accuracy. 
% Additionally, we have shown that using a non-network based sun glare detection algorithm to split images between a sun glare trained detector and a detector trained on normal conditions leads to improved results (we haven't done this yet).

\vspace{-3mm}

The GLARE dataset is the beginning of future research on traffic sign detection in naturally noisy conditions and the removal of sun glare as well. We believe this dataset can be used as a testing set for entire sun glare removal using the U-Net architecture~\cite{unet}, as seen in previous work removing sun flares from images~\cite{sunflare}. We also believe this dataset can be extended to include traffic signs in other noisy or abnormal conditions, such as rain, fog, and night-time driving. Such an extension could be used to create and train architectures that can detect traffic signs with greater precision in a wider variety of conditions, whether through image restoration or detection and recognition alone. {Finally, we believe this dataset can be used to assist in the development of object detection architectures that can detect objects well despite localized noise, as possibly evidenced by TOOD.}

\vspace{-5mm}
\bibliographystyle{unsrt}
\bibliography{main}

\begin{thebibliography}{10}

\bibitem{DoTreport}
National Highway Traffic Safety~Administration U.S. Department~of Transportation.
\newblock Traffic safety facts: Crash stats --- sun glare and slick roads are the two most environment-related causing circumstances., 2015.

\bibitem{bian2022machine}
Jiang Bian, Abdullah Al~Arafat, Haoyi Xiong, Jing Li, Li~Li, Hongyang Chen, Jun Wang, Dejing Dou, and Zhishan Guo.
\newblock Machine learning in real-time internet of things (iot) systems: A survey.
\newblock {\em IEEE Internet of Things Journal}, 9(11):8364--8386, 2022.

\bibitem{mogelmose2012vision}
Andreas Mogelmose, Mohan~Manubhai Trivedi, and Thomas~B Moeslund.
\newblock Vision-based traffic sign detection and analysis for intelligent driver assistance systems: Perspectives and survey.
\newblock {\em IEEE Transactions on Intelligent Transportation Systems}, 13(4):1484--1497, 2012.

\bibitem{larsson2011correlating}
Fredrik Larsson, Michael Felsberg, and P-E Forssen.
\newblock Correlating fourier descriptors of local patches for road sign recognition.
\newblock {\em IET Computer Vision}, 5(4):244--254, 2011.

\bibitem{stallkamp2012man}
Johannes Stallkamp, Marc Schlipsing, Jan Salmen, and Christian Igel.
\newblock Man vs. computer: Benchmarking machine learning algorithms for traffic sign recognition.
\newblock {\em Neural networks}, 32:323--332, 2012.

\bibitem{houben2013detection}
Sebastian Houben, Johannes Stallkamp, Jan Salmen, Marc Schlipsing, and Christian Igel.
\newblock Detection of traffic signs in real-world images: The german traffic sign detection benchmark.
\newblock In {\em The 2013 international joint conference on neural networks (IJCNN)}, pages 1--8. Ieee, 2013.

\bibitem{zhu2016traffic}
Zhe Zhu, Dun Liang, Songhai Zhang, Xiaolei Huang, Baoli Li, and Shimin Hu.
\newblock Traffic-sign detection and classification in the wild.
\newblock In {\em Proceedings of the IEEE conference on computer vision and pattern recognition}, pages 2110--2118, 2016.

\bibitem{ertler2020mapillary}
Christian Ertler, Jerneja Mislej, Tobias Ollmann, Lorenzo Porzi, Gerhard Neuhold, and Yubin Kuang.
\newblock The mapillary traffic sign dataset for detection and classification on a global scale.
\newblock In {\em European Conference on Computer Vision}, pages 68--84. Springer, 2020.

\bibitem{tabernik2019deep}
Domen Tabernik and Danijel Sko{\v{c}}aj.
\newblock Deep learning for large-scale traffic-sign detection and recognition.
\newblock {\em IEEE transactions on intelligent transportation systems}, 21(4):1427--1440, 2019.

\bibitem{timofte2014multi}
Radu Timofte, Karel Zimmermann, and Luc Van~Gool.
\newblock Multi-view traffic sign detection, recognition, and 3d localisation.
\newblock {\em Machine vision and applications}, 25(3):633--647, 2014.

\bibitem{temel2019traffic}
Dogancan Temel, Min-Hung Chen, and Ghassan AlRegib.
\newblock Traffic sign detection under challenging conditions: A deeper look into performance variations and spectral characteristics.
\newblock {\em IEEE Transactions on Intelligent Transportation Systems}, 21(9):3663--3673, 2019.

\bibitem{soendoro2011traffic}
David Soendoro and Iping Supriana.
\newblock Traffic sign recognition with color-based method, shape-arc estimation and svm.
\newblock In {\em Proceedings of the 2011 International Conference on Electrical Engineering and Informatics}, pages 1--6. IEEE, 2011.

\bibitem{mao2016hierarchical}
Xuehong Mao, Samer Hijazi, Ra{\'u}l Casas, Piyush Kaul, Rishi Kumar, and Chris Rowen.
\newblock Hierarchical cnn for traffic sign recognition.
\newblock In {\em 2016 IEEE intelligent vehicles symposium (IV)}, pages 130--135. IEEE, 2016.

\bibitem{arcos2018deep}
{\'A}lvaro Arcos-Garc{\'\i}a, Juan~A Alvarez-Garcia, and Luis~M Soria-Morillo.
\newblock Deep neural network for traffic sign recognition systems: An analysis of spatial transformers and stochastic optimisation methods.
\newblock {\em Neural Networks}, 99:158--165, 2018.

\bibitem{krizhevsky2012imagenet}
Alex Krizhevsky, Ilya Sutskever, and Geoffrey~E Hinton.
\newblock Imagenet classification with deep convolutional neural networks.
\newblock {\em Advances in neural information processing systems}, 25, 2012.

\bibitem{girshick2014rich}
Ross Girshick, Jeff Donahue, Trevor Darrell, and Jitendra Malik.
\newblock Rich feature hierarchies for accurate object detection and semantic segmentation.
\newblock In {\em Proceedings of the IEEE conference on computer vision and pattern recognition}, pages 580--587, 2014.

\bibitem{girshick2015fast}
Ross Girshick.
\newblock Fast r-cnn.
\newblock In {\em Proceedings of the IEEE international conference on computer vision}, pages 1440--1448, 2015.

\bibitem{redmon2016you}
Joseph Redmon, Santosh Divvala, Ross Girshick, and Ali Farhadi.
\newblock You only look once: Unified, real-time object detection.
\newblock In {\em Proceedings of the IEEE conference on computer vision and pattern recognition}, pages 779--788, 2016.

\bibitem{ge2021yolox}
Zheng Ge, Songtao Liu, Feng Wang, Zeming Li, and Jian Sun.
\newblock Yolox: Exceeding yolo series in 2021.
\newblock {\em arXiv preprint arXiv:2107.08430}, 2021.

\bibitem{carion2020end}
Nicolas Carion, Francisco Massa, Gabriel Synnaeve, Nicolas Usunier, Alexander Kirillov, and Sergey Zagoruyko.
\newblock End-to-end object detection with transformers.
\newblock In {\em European conference on computer vision}, pages 213--229. Springer, 2020.

\bibitem{liu2021swin}
Ze~Liu, Yutong Lin, Yue Cao, Han Hu, Yixuan Wei, Zheng Zhang, Stephen Lin, and Baining Guo.
\newblock Swin transformer: Hierarchical vision transformer using shifted windows.
\newblock In {\em Proceedings of the IEEE/CVF International Conference on Computer Vision}, pages 10012--10022, 2021.

\bibitem{videolabel}
Nathan de~Bruijn.
\newblock https://github.com/natdebru/opencv-video-label, 2019.

\bibitem{gordon2018re3}
Daniel Gordon, Ali Farhadi, and Dieter Fox.
\newblock Re3: Real-time recurrent regression networks for visual tracking of generic objects.
\newblock {\em IEEE Robotics and Automation Letters}, 3(2):788--795, 2018.

\bibitem{Nebehay2015CVPR}
Georg Nebehay and Roman Pflugfelder.
\newblock Clustering of {Static-Adaptive} correspondences for deformable object tracking.
\newblock In {\em Computer Vision and Pattern Recognition}. IEEE, June 2015.

\bibitem{lin2014microsoft}
Tsung-Yi Lin, Michael Maire, Serge Belongie, James Hays, Pietro Perona, Deva Ramanan, Piotr Dollar, and Larry Zitnick.
\newblock Microsoft coco: Common objects in context.
\newblock In {\em ECCV}. European Conference on Computer Vision, September 2014.

\bibitem{glenn_jocher_2022_6222936}
Glenn Jocher, Ayush Chaurasia, Alex Stoken, Jirka Borovec, NanoCode012, Yonghye Kwon, TaoXie, Jiacong Fang, imyhxy, Kalen Michael, Lorna, Abhiram V, Diego Montes, Jebastin Nadar, Laughing, tkianai, yxNONG, Piotr Skalski, Zhiqiang Wang, Adam Hogan, Cristi Fati, Lorenzo Mammana, AlexWang1900, Deep Patel, Ding Yiwei, Felix You, Jan Hajek, Laurentiu Diaconu, and Mai~Thanh Minh.
\newblock {ultralytics/yolov5: v6.1 - TensorRT, TensorFlow Edge TPU and OpenVINO Export and Inference}, February 2022.

\bibitem{Jocher_YOLO_by_Ultralytics_2023}
Glenn Jocher, Ayush Chaurasia, and Jing Qiu.
\newblock {YOLO by Ultralytics}, 1 2023.

\bibitem{mmdetection}
Kai Chen, Jiaqi Wang, Jiangmiao Pang, Yuhang Cao, Yu~Xiong, Xiaoxiao Li, Shuyang Sun, Wansen Feng, Ziwei Liu, Jiarui Xu, Zheng Zhang, Dazhi Cheng, Chenchen Zhu, Tianheng Cheng, Qijie Zhao, Buyu Li, Xin Lu, Rui Zhu, Yue Wu, Jifeng Dai, Jingdong Wang, Jianping Shi, Wanli Ouyang, Chen~Change Loy, and Dahua Lin.
\newblock {MMDetection}: Open mmlab detection toolbox and benchmark.
\newblock {\em arXiv preprint arXiv:1906.07155}, 2019.

\bibitem{feng2021tood}
Chengjian Feng, Yujie Zhong, Yu~Gao, Matthew~R Scott, and Weilin Huang.
\newblock Tood: Task-aligned one-stage object detection.
\newblock In {\em ICCV}, 2021.

\bibitem{zhu2021deformable}
Xizhou Zhu, Weijie Su, Lewei Lu, Bin Li, Xiaogang Wang, and Jifeng Dai.
\newblock Deformable detr: Deformable transformers for end-to-end object detection.
\newblock In {\em International Conference on Learning Representations}, 2021.

\bibitem{Ren_2017}
Shaoqing Ren, Kaiming He, Ross Girshick, and Jian Sun.
\newblock Faster r-cnn: Towards real-time object detection with region proposal networks.
\newblock {\em IEEE Transactions on Pattern Analysis and Machine Intelligence}, Jun 2017.

\bibitem{resnet}
Kaiming He, Xiangyu Zhang, Shaoqing Ren, and Jian Sun.
\newblock Deep residual learning for image recognition, 2015.

\bibitem{Jocher_YOLOv5_by_Ultralytics_2020}
Glenn Jocher.
\newblock {YOLOv5 by Ultralytics}, 5 2020.

\bibitem{deng2009imagenet}
Jia Deng, Wei Dong, Richard Socher, Li-Jia Li, Kai Li, and Li~Fei-Fei.
\newblock Imagenet: A large-scale hierarchical image database.
\newblock In {\em 2009 IEEE conference on computer vision and pattern recognition}, pages 248--255. Ieee, 2009.

\bibitem{unet}
Olaf Ronneberger, Philipp Fischer, and Thomas Brox.
\newblock U-net: Convolutional networks for biomedical image segmentation, 2015.

\bibitem{sunflare}
Yicheng Wu, Qiurui He, Tianfan Xue, Rahul Garg, Jiawen Chen, Ashok Veeraraghavan, and Jonathan~T. Barron.
\newblock How to train neural networks for flare removal, 2020.

\end{thebibliography}

\end{document}